\documentclass[11pt,a4paper]{article}
\usepackage[nohyperref]{naaclhlt2018}
\usepackage{times}
\usepackage{latexsym}

\usepackage{url}

\usepackage{blindtext}
\usepackage{latexsym}
\usepackage{multirow}
\usepackage{amsfonts}
\usepackage{amssymb}
\usepackage{amsmath}
\usepackage{amsthm}
\usepackage{setspace}
\usepackage{enumitem}
\usepackage{tikz}
\usepackage{calc}
\usetikzlibrary{shapes}
\usepackage{array}
\usepackage{xcolor}
\usepackage{booktabs}
\def\mybar#1#2#3{
  #1 & {\color{gray}\rule{#1pt*\real{#3}-#2pt*\real{#3}}{8pt}}}
\def\bestbar#1#2#3{
  \textbf{#1} & {\color{black}\rule{#1pt*\real{#3}-#2pt*\real{#3}}{8pt}}}

\DeclareMathOperator{\softmax}{softmax}

\theoremstyle{plain}
\newtheorem*{conjecture*}{Conjecture}

\setlist{nosep}
\setlist[itemize]{leftmargin=*}

\makeatletter
\g@addto@macro\normalsize{%
  \setlength\abovedisplayskip{5pt}
  \setlength\belowdisplayskip{5pt}
  \setlength\abovedisplayshortskip{5pt}
  \setlength\belowdisplayshortskip{5pt}
}
\makeatother

\aclfinalcopy 

\title{Reusing Weights in Subword-aware Neural Language Models}

\author{Zhenisbek Assylbekov \\
  School of Science and Technology \\
  Nazarbayev University \\
  {\tt zhassylbekov@nu.edu.kz} \\\And
  Rustem Takhanov \\
  School of Science and Technology \\
  Nazarbayev University \\
  {\tt rustem.takhanov@nu.edu.kz} \\}

\date{}

\begin{document}
\maketitle
\begin{abstract}
We propose several ways of reusing  subword embeddings and other weights in subword-aware neural language models. The proposed techniques do not benefit a competitive character-aware model, but some of them improve the performance of syllable- and morpheme-aware models while showing significant reductions in model sizes. We discover a simple hands-on principle: in a multi-layer input embedding model, layers should be tied consecutively bottom-up if reused at output. Our best morpheme-aware model with properly reused weights beats the competitive word-level model by a large margin across multiple languages and has 20\%--87\% fewer parameters. 
\end{abstract}

\section{Introduction}
A statistical language model (LM) is a model which assigns a probability to a sequence of words. 
It is used in speech recognition, 
machine translation, 
part-of-speech tagging, 
information retrieval 
and other applications. Data sparsity is a major problem in building traditional $n$-gram language models, which assume that the probability of a word only depends on the previous $n$ words. To deal with potentially severe problems when confronted with any $n$-grams that have not explicitly been seen before, some form of smoothing is necessary.

Recent progress in statistical language modeling is connected with neural language models (NLM), which tackle the data sparsity problem by representing words as vectors. Typically this is done twice: at input (to embed the current word of a sequence into a vector space) and at output (to embed candidates for the next word of a sequence). Especially successful are the models in which the architecture of the neural network between input and output is recurrent \cite{mikolov2010recurrent}, which we refer to as recurrent neural network language models (RNNLM). 

Tying input and output word embeddings in word-level RNNLM  is a regularization technique, which was introduced earlier \cite{bengio2001neural,mnih2007three} but has been widely used relatively recently, and there is  empirical evidence \cite{press2016using} as well as theoretical justification \cite{inan2016tying} that such a simple trick improves  language modeling quality while decreasing the total number of trainable parameters almost two-fold, since most of the parameters are due to embedding matrices. Unfortunately, this regularization technique is not directly applicable to subword-aware neural language models as they receive subwords at input and return words at output. This raises the following questions: Is it possible to reuse embeddings and other parameters in subword-aware neural language models? Would it benefit language modeling quality? We experimented with different subword units, embedding models, and ways of reusing parameters, and our answer to both questions is as follows: There are several ways to reuse weights in subword-aware neural language models, and none of them improve a competitive character-aware model, but some of them do benefit syllable- and morpheme-aware models, while giving significant reductions in model sizes. A simple morpheme-aware model that sums morpheme embeddings of a word benefits most from appropriate weight tying, showing a significant gain over the competitive word-level baseline across different languages and data set sizes.
Another contribution of this paper is the discovery of a hands-on principle that in a multi-layer input embedding model, layers should be tied consecutively bottom-up if reused at output. 

The source code for the morpheme-aware model is available at \url{https://github.com/zh3nis/morph-sum}.

\section{Related Work}
\noindent\textbf{Subword-aware NLM:} There has been a large number of publications in the last 2--3 years on subword-level and subword-aware NLMs,\footnote{\textit{Subword-level} LMs rely on subword-level inputs and make predictions at the level of subwords; \textit{subword-aware} LMs also rely on subword-level inputs but make predictions at the level of words.} especially for the cases when subwords are characters \cite{ling-EtAl:2015:EMNLP2,kim2016character,verwimp2017character} or morphemes \cite{botha2014compositional,qiu2014co,cotterell2015morphological}. Less work has been done on syllable-level or syllable-aware NLMs \cite{mikolov2012subword,assylbekov2017syllable,yu2017syllable}. For a thorough and up-to-date review of the previous work on subword-aware neural language modeling we refer the reader to the paper by \citeauthor{vania2017characters} \shortcite{vania2017characters}, where the authors systematically compare different subword units (characters, character trigrams, BPE, morphs/morphemes) and different representation models (CNN, Bi-LSTM, summation) on languages with various morphological typology. 

\noindent\textbf{Tying weights in NLM:} Reusing embeddings in word-level neural language models is a technique which was 
used earlier \cite{bengio2001neural,mnih2007three} and studied in more details recently 
\cite{inan2016tying,press2016using}. However, not much work has been done on reusing parameters in subword-aware or subword-level language models. \citeauthor{jozefowicz2016exploring} \shortcite{jozefowicz2016exploring} reused the {\it CharCNN} architecture of \citeauthor{kim2016character} \shortcite{kim2016character} to dynamically generate softmax word embeddings without sharing  parameters with an input word-embedding sub-network. They managed to significantly reduce the total number of parameters for large models trained on a huge dataset in English (1B tokens) with a large vocabulary (800K tokens) at the expense of deteriorated performance. \citet{labeau2017character} used similar approach to augment the output word representations with subword-based embeddings. They experimented with characters and morphological decompositions, and tried different compositional models (CNN, Bi-LSTM, concatenation) on Czech dataset consisting of 4.7M tokens. They were not tying weights  between input and output
representations, since their preliminary experiments with tied weights gave worse results. 

Our approach differs in the following aspects: we focus on the ways to {\it reuse} weights at output, seek both model size reduction \textit{and} performance improvement in subword-aware language models, try different subword units (characters, syllables, and morphemes), and make evaluation on small (1M--2M tokens) and medium (17M--51M tokens) data sets across multiple languages.

\section{Recurrent Neural Language Model}
Let $\mathcal{W}$ be a finite vocabulary of words. We assume that words have already been converted into indices. Let $\mathbf{E_\mathcal{W}^\text{in}}\in\mathbb{R}^{|\mathcal{W}|\times d_\mathcal{W}}$ be an input embedding matrix for words --- i.e., it is a matrix in which the $w$th row (denoted as $\mathbf{w}$) corresponds to an embedding of the word $w\in \mathcal{W}$.

Based on word embeddings $\mathbf{w_{1:k}}=\mathbf{w_1},\ldots,\mathbf{w_k}$ for a sequence of words $w_{1:k}$, a typical word-level RNN language model produces a sequence of states $\mathbf{h_{1:k}}$ according to
\begin{align}
\mathbf{h_t} = \text{RNNCell}(\mathbf{w_t}, \mathbf{h_{t-1}}),\quad\mathbf{h_0} = \mathbf{0}.\label{seq_model}
\end{align}
The last state $\mathbf{h_k}$ is assumed to contain information on the whole sequence $w_{1:k}$ and is further used for predicting the next word $w_{k+1}$ of a sequence according to the probability distribution
\begin{equation}
\Pr(w_{k+1}|w_{1:k})=\softmax(\mathbf{h_k}\mathbf{E}_\mathcal{W}^\text{out} + \mathbf{b}),\label{softmax}
\end{equation}
where $\mathbf{E}_\mathcal{W}^\text{out}\in\mathbb{R}^{d_{\text{LM}}\times|\mathcal{W}|}$ is an output embedding matrix, $\mathbf{b}\in\mathbb{R}^{|\mathcal{W}|}$ is a bias term, and $d_{\text{LM}}$ is a state size of the RNN.

\noindent\textbf{Subword-based word embeddings:}
One of the more recent advancements in neural language modeling has to do with segmenting words at input into subword units (such as characters, syllables, morphemes, etc.) and composing each word's embedding from the embeddings of its subwords. Formally, let $\mathcal{S}$ be a finite vocabulary of subwords,\footnote{As in the case of words, we  assume that subwords have already been converted into indices.} and let $\mathbf{E}_{\mathcal{S}}^\text{in}\in\mathbb{R}^{|\mathcal{S}|\times d_\mathcal{S}}$ be an input embedding matrix for subwords. Any word $w\in\mathcal{W}$ is a sequence of its subwords $(s_1, s_2, \ldots, s_{n_w}) = \sigma(w)$, and hence can be represented as a sequence of the corresponding subword vectors:
\begin{equation}
[\mathbf{s_1}, \mathbf{s_2}, \ldots, \mathbf{s_{n_w}}].\label{subw_seq}
\end{equation} 
A subword-based word embedding model $E(\cdot; \mathbf{E}_\mathcal{S}^\text{in}, \boldsymbol{\Theta}^\text{in})$ with parameters $(\mathbf{E}_\mathcal{S}^\text{in}, \boldsymbol{\Theta}^\text{in})$ constructs a word vector $\mathbf{x}$ from the sequence of subword vectors (\ref{subw_seq}), i.e.
\begin{equation}
\mathbf{x} = E(\sigma(w); \mathbf{E}_\mathcal{S}^\text{in},\boldsymbol{\Theta}^\text{in}),\label{emb_model}
\end{equation}
which is then fed into a RNNLM (\ref{seq_model}) instead of a plain embedding $\mathbf{w}$. The additional parameters $\boldsymbol{\Theta^\text{in}}$ correspond to the way the embedding model constructs the word vector: for instance, in the {\it CharCNN} model of \citeauthor{kim2016character} \shortcite{kim2016character}, $\boldsymbol{\Theta^\text{in}}$ are the weights of the convolutional and highway layers.

\noindent\textbf{Reusing word embeddings:} Another recent technique in word-level neural language modeling is tying input and output word embeddings:
$$
\mathbf{E}_\mathcal{W}^\text{in}=\left(\mathbf{E}_\mathcal{W}^\text{out}\right)^T,
$$
under the assumption that $d_\mathcal{W}=d_\text{LM}$. Although being useful for word-level language modeling \cite{press2016using,inan2016tying}, this regularization technique is not directly applicable to subword-aware language models, as they receive subword embeddings at input and return word embeddings at output. In the next section we describe a simple technique to allow reusing subword embeddings $\mathbf{E}^\text{in}_\mathcal{S}$ as well as other parameters $\mathbf{\Theta}^\text{in}$ in a subword-aware RNNLM.

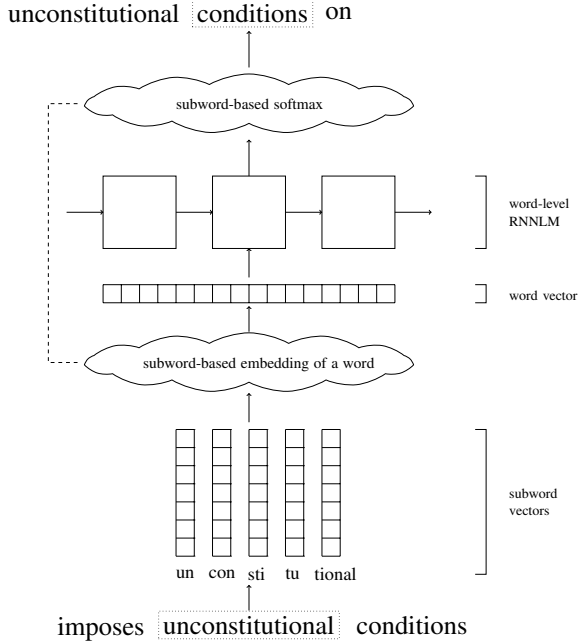
\begin{figure}
  \scalebox{.48}{
    \begin{tikzpicture}
      \node[text width=7cm] at (-0.1,3.5) {\huge unconstitutional}; 
      \node[draw,dotted] at (3.1,3.5) {\huge conditions};
      \node[text width=5cm] at (7.6,3.5) {\huge on};
      \draw[->] (3,2) -- (3,3);
      \node [cloud, draw,cloud puffs=13,cloud puff arc=60, aspect=8.35, inner ysep=1em] at (3, 1.0) {};
      \node[text width=10cm] at (6, 1.0)
    {\large subword-based softmax};

      \draw[->] (3,-1) -- (3,0);

      \draw[->] (-2,-2) -- (-1,-2);
      \draw [black] (-1,-3) rectangle (1,-1);
      [line width=20pt]
      \draw[->] (1,-2) -- (2,-2);
      \draw [black] (2,-3) rectangle (4,-1);
      \draw[->] (4,-2) -- (5,-2);
      \draw [black] (5,-3) rectangle (7,-1);
      \draw[->] (7,-2) -- (8,-2);
      \draw (9.2, -3) -- (9.5,-3) -- (9.5,-1) -- (9.2, -1);
      
      \node[text width=2.0cm, anchor=west, right] at (10,-2)
    {word-level RNNLM};
      
      \draw[->] (3,-3.8) -- (3,-3);
      \draw[step=0.5 cm,black,very thin] (-1,-4) grid (7,-4.5);
      
      \draw[->] (3,-5.3) -- (3,-4.6);
      
      \draw (9.2, -4.5) -- (9.5,-4.5) -- (9.5,-4) -- (9.2, -4);
      \node[text width=7cm, anchor=west, right] at (10,-4.3)
    {word vector};
    
      \node [cloud, draw,cloud puffs=13,cloud puff arc=60, aspect=8.35, inner ysep=1em] at (3, -6.15) {};
      
      \draw[dashed] (-1.7, 1) -- (-2.5, 1) -- (-2.5,-6.15) -- (-1.7, -6.15);
      
      \node[text width=10cm] at (5.1,-6.2)
    {\large subword-based embedding of a word};
      
      \draw[->] (3,-7.8) -- (3,-7);
      
      \draw (1,-8) -- (1,-11.5);
      \draw[step=0.5 cm,black,very thin] (1.5,-8) grid (1,-11.5);
      
      \draw[step=0.5 cm,very thin] (1.5,-8) grid (1.5,-11.5);
      \draw[step= 0.5 cm,very thin] (2,-8) grid (2.5,-11.5);
      
      \draw[step= 0.5 cm,very thin] (2.5,-8) grid (2.5,-11.5);
      \draw[step=0.5 cm,very thin] (3,-8) grid (3.5,-11.5);
      
      \draw[step=0.5 cm,very thin] (3.5,-8) grid (3.5,-11.5);
      \draw[step=0.5 cm,very thin] (4,-8) grid (4.5,-11.5);
      
      \draw[step=0.5 cm,very thin] (5,-8) grid (5.5,-11.5);
      \draw[step=0.5 cm,very thin] (4.5,-8) grid (4.5,-11.5);
      
      \draw (9.2, -8) -- (9.5,-8) -- (9.5,-12) -- (9.2, -12);
      \node[text width=1.9cm, anchor=west, right] at (10,-10)
    {subword vectors};
      
      \node[text width=2cm] at (2,-12) {\Large un};
      \node[text width=3cm] at (3.4,-12) {\Large con};
      \node[text width=3cm] at (4.47,-12) {\Large sti};
      \node[text width=3cm] at (5.5,-12) {\Large tu};
      \node[text width=3cm] at (6.3,-12) {\Large tional};
      
      \draw[->] (3,-13) -- (3,-12.3);
      
      \node[text width=7cm] at (1.25,-13.5) {\huge imposes};
      \node[draw,dotted] at (3,-13.4) {\huge unconstitutional};
      \node[text width=4cm] at (7.95,-13.4) {\huge conditions};
    \end{tikzpicture}
  }
  \caption{Subword-aware RNNLM with subword-based softmax.}
  \label{architecture}
\end{figure}

\section{Reusing Weights}
\label{re_emb_section}
Let $\mathbf{E}_\mathcal{S}^\text{out}$ be an output embedding matrix for subwords and let us modify the softmax layer (\ref{softmax}) so that it utilizes $\mathbf{E}_\mathcal{S}^\text{out}$ instead of the word embedding matrix $\mathbf{E}_\mathcal{W}^\text{out}$. The idea is fairly straightforward: we reuse an embedding model (\ref{emb_model}) to construct a new word embedding matrix:
\begin{equation}
\mathbf{\hat{E}}_\mathcal{W}^\text{out}=[E(\sigma(w); \mathbf{E}_\mathcal{S}^\text{out},\boldsymbol{\Theta}^\text{out})\text{ for }w\in\mathcal{W}],\label{new_emb}
\end{equation}
and use $\mathbf{\hat{E}}_\mathcal{W}^\text{out}$ instead of $\mathbf{{E}}_\mathcal{W}^\text{out}$ in the softmax layer (\ref{softmax}). Such modification of the softmax layer will be referred to as \textit{subword-based softmax}. The overall architecture of a subword-aware  RNNLM with subword-based softmax is given in Figure \ref{architecture}. Such a model allows several options for reusing embeddings and weights, which are discussed below.
\begin{itemize}
\item{Reusing neither subword embeddings nor embedding model weights:} As was shown by \citeauthor{jozefowicz2016exploring} \shortcite{jozefowicz2015empirical}, this can significantly reduce the total number of parameters for large models trained on huge datasets (1B tokens) with large vocabularies (800K tokens). However, we do not expect significant reductions on smaller data sets (1-2M tokens) with smaller vocabularies (10-30K tokens), which we use in our main experiments.

\item{Reusing subword embeddings (\textit{RE})} can be done by setting $\mathbf{E}_\mathcal{S}^\text{out}=\mathbf{E}_\mathcal{S}^\text{in}$ in (\ref{new_emb}). This will give a significant reduction in model size for models with $|\mathbf{E^\text{in}_\mathcal{S}}|\gg|\boldsymbol{\Theta}^\text{in}|$,\footnote{$|\mathbf{A}|$ denotes number of elements in $\mathbf{A}$.} such as the morpheme-aware model of \citeauthor{botha2014compositional} \shortcite{botha2014compositional}.

\item{Reusing weights of the embedding model (\textit{RW})} can be done by setting $\boldsymbol{\Theta}^\text{out}=\boldsymbol{\Theta}^\text{in}$. Unlike the previous option, this should significantly reduce sizes of models with $|\mathbf{E^\text{in}_\mathcal{S}}|\ll|\boldsymbol{\Theta}^\text{in}|$, such as the character-aware model of \citeauthor{kim2016character} \shortcite{kim2016character}.

\item{Reusing both subword embeddings and weights of the embedding model (\textit{RE+RW})} can be done by setting
$\mathbf{E}_\mathcal{S}^\text{out}=\mathbf{E}_\mathcal{S}^\text{in}$ and $\boldsymbol{\Theta}^\text{out}=\boldsymbol{\Theta}^\text{in}$ simultaneously in (\ref{new_emb}). This should significantly reduce the number of trainable parameters in any subword-aware model. Here we use exactly the same word representations both at input and at output, so this option corresponds to the reusing of plain word embeddings in pure word-level language models.
\end{itemize}

\section{Experimental Setup}\label{exp_setup}
\textbf{Data sets:} All models are trained and evaluated on the PTB \cite{marcus1993building} and the WikiText-2 \cite{merity2016pointer} data sets. For the PTB we utilize the standard training (0-20), validation (21-22), and test (23-24) splits along with pre-processing per \citeauthor{mikolov2010recurrent} \shortcite{mikolov2010recurrent}. WikiText-2 is an alternative to PTB, which is approximately two times as large in size and three times as large in vocabulary (Table~\ref{corpus_stats}).

\begin{table}[h]
\begin{center}
\begin{tabular}{l r r r r}
\toprule
Data set & \multicolumn{1}{c}{$T$} & \multicolumn{1}{c}{$|\mathcal{W}|$} & \multicolumn{1}{c}{$|\mathcal{S}|$} & \multicolumn{1}{c}{$|\mathcal{M}|$}\\
 \midrule
PTB & 0.9M & 10K & 5.9K & 3.4K\\
WikiText-2 & 2.1M & 33K & 19.5K & 8.8K\\
\bottomrule
\end{tabular}
\end{center}
\vspace{-10pt}\caption{Corpus statistics. $T=$ number of tokens in training set; $|\mathcal{W}|=$ word vocabulary size; $|\mathcal{S}|=$ syllable vocabulary size; $|\mathcal{M}|=$ morph vocabulary size.}
\label{corpus_stats}
\end{table}

\noindent\textbf{Subword-based embedding models:}
We experiment with existing representational models which have previously proven effective for language modeling.
\begin{itemize}
\item\textit{CharCNN} \cite{kim2016character} is a character-aware convolutional model, which performs on par with the 2014--2015 state-of-the-art word-level LSTM model \cite{zaremba2014recurrent} despite having 60\% fewer parameters.

\item\textit{SylConcat} is a simple concatenation of syllable embeddings suggested by \citeauthor{assylbekov2017syllable} \shortcite{assylbekov2017syllable}, which underperforms {\it CharCNN} but has fewer parameters and is trained faster.
\item\textit{MorphSum} is a summation of morpheme embeddings, which is similar to the approach of \citeauthor{botha2014compositional} \shortcite{botha2014compositional} with one important difference: the embedding of the word itself is not included into the sum. We do this since other models do not utilize word embeddings. 
\end{itemize}
In all subword-aware language models we inject a stack of two highway layers \cite{srivastava2015training} right before the word-level RNNLM as done by \citeauthor{kim2016character} \shortcite{kim2016character}, and the non-linear activation in any of these highway layers is a ReLU. The highway layer size is denoted by $d_\text{HW}$.

\begin{table*}[h]
\begin{center}
\begin{tabular}{l | c c | c c | c c | c c}
\toprule
 & \multicolumn{4}{c|}{PTB}  & \multicolumn{4}{c}{Wikitext-2} \\
\midrule
\multirow{2}{*}{Model} & \multicolumn{2}{c|}{Small} & \multicolumn{2}{c|}{Medium} &  \multicolumn{2}{c|}{Small} & \multicolumn{2}{c}{Medium}\\
 & Size & PPL & Size & PPL & Size & PPL & Size & PPL \\
\midrule
Word                       & 4.7M & 88.1 & 19.8M & 79.8 & 14M & 111.9 & 50.1M & 95.7 \\
Word + reusing word emb's  & 2.7M & {86.6} & 13.3M & 74.5 & 7.3M & {104.1} & 28.4M & 89.9 \\
\midrule
\midrule
CharCNN (original)  & 4.1M & \textbf{87.3} & 19.4M & \textbf{77.1} & 8.7M & \textbf{101.6} & 34.5M & 88.7 \\
CharCNN               & 3.3M & 97.5 & 18.5M & 89.2 & 3.3M & 110.6 & 18.5M & --- \\
CharCNN + RE         & 3.3M & 99.1 & 18.5M & 82.9 & 3.3M & 110.2 & 18.5M & --- \\
CharCNN + RW         & 2.2M & 93.5 & 13.6M & 103.2 & 2.2M & 111.5 & 13.6M & --- \\
CharCNN + RE + RW    & 2.2M & 91.0 & 13.6M & 79.9 & 2.2M & 101.8 & 13.6M & --- \\
\midrule
SylConcat (original) & 3.2M & 89.0 & 18.7M & 77.9 & 8.5M & 105.7 & 36.6M &  91.4\\
SylConcat            & 1.7M & 96.9 & 17.7M & 90.5 & 3.1M & 118.1 & 23.2M & 114.8 \\
SylConcat + RE       & 1.4M & \textbf{87.4} & 16.6M & \textbf{75.7} & 2.1M & \textbf{101.0} & 19.3M &  94.2\\
SylConcat + RW       & 1.6M & 99.9 & 15.2M & 96.2 & 2.9M & 118.9 & 19.4M & 112.1\\
SylConcat + RE + RW & 1.2M & 88.4 & 12.7M & 76.2 & 1.9M & \textbf{101.0} & 15.5M& \textbf{86.7} \\
\midrule
MorphSum (original)      & 3.5M & 87.5 & 17.2M & 78.5 & 9.3M & 101.9 & 35.8M & 90.1\\
MorphSum             & 2.4M & 89.0 & 14.5M & 82.4 & 4.5M & 100.3 & 21.7M & 86.7 \\
MorphSum + RE        & 1.6M & 85.5 & 12.3M & 74.1 & 2.8M & 97.6 & 15.9M & 81.2 \\
MorphSum + RW        & 2.2M & 89.6 & 12.8M & 81.0 & 4.4M & 101.4 & 20.0M & 86.6 \\
MorphSum + RE + RW   & 1.5M & \textbf{85.1} & 10.7M & \textbf{72.2} & 2.6M & \textbf{96.5} & 14.2M & \textbf{77.5}\\
\midrule
\end{tabular}
\end{center}
\vspace{-10pt}\caption{Results. The pure word-level models  and original versions of subword-aware models (with regular softmax) serve as baselines. Reusing the input embedding architecture at output in {\it CharCNN} leads to prohibitively slow models when trained on WikiText-2 ($\approx$800 tokens/sec on NVIDIA Titan X Pascal); we therefore abandoned evaluation of these configurations.}
\label{results}
\end{table*}

\noindent\textbf{Word-level RNNLM:} There is a large variety of RNN cells to choose from in (\ref{seq_model}). To make our results directly comparable to the previous work of \citeauthor{inan2016tying} \shortcite{inan2016tying}, \citeauthor{press2016using} \shortcite{press2016using} on reusing word embeddings we select a rather conventional architecture -- a stack of two LSTM cells \cite{hochreiter1997long}.

\noindent\textbf{Hyperparameters:} We experiment with two configurations for the state size $d_\text{LM}$ of the word-level RNNLM: 200 (small models) and 650 (medium-sized models). In what follows values outside brackets correspond to small models, and values within brackets correspond to medium models.
\begin{itemize}
\item\textit{CharCNN:} We use the same hyperparameters as in the work of \citeauthor{kim2016character} \shortcite{kim2016character}, where ``large model'' stands for what we call ``medium-sized model''.


\item\textit{SylConcat:} $d_\mathcal{S}=50$ ($200$), $d_\text{HW}=200$ ($800$). These choices are guided by the work of \citeauthor{assylbekov2017syllable} \shortcite{assylbekov2017syllable}.

\item\textit{MorphSum}: $d_\mathcal{S}=d_\text{HW}=200$ ($650$). These choices are guided by \citeauthor{kim2016character} \shortcite{kim2016character}.
\end{itemize}

\noindent\textbf{Optimizaton} method is guided by the previous works \cite{zaremba2014recurrent,gal2016theoretically} on word-level language modeling with LSTMs. See Appendix 
\ref{opt} for details. 

\noindent\textbf{Syllabification and morphological segmentation:} True syllabification of a word requires its grapheme-to-phoneme conversion and then its splitting up into syllables based on some rules. True morphological segmentation requires rather expensive morphological analysis and disambiguation tools. Since these are not always available for under-resourced languages, we decided to utilize Liang's widely-used hyphenation algorithm \cite{liang1983word} and an unsupervised morphological segmentation tool,
Morfessor 2.0 \cite{virpioja2013morfessor}, as approximations to syllabification and morphological segmentation respectively. We use the default configuration of Morfessor 2.0. Syllable and morpheme vocabulary sizes for both PTB and WikiText-2 are reported in Table~\ref{corpus_stats}.

\section{Results}
In order to investigate the extent to which each of our proposed options benefits the language modeling task,
we evaluate all four modifications (no reusing, RE, RW, RE+RW) for each subword-aware model against their original versions and word-level baselines. The results of evaluation are given in Table \ref{results}. We have both negative and positive findings which are summarized below.
\vspace{5pt}

\noindent\textbf{Negative results:}
\begin{itemize}
\item The `no reusing' and RW options should never be applied in subword-aware language models as they deteriorate the performance.
\item Neither of the reusing options benefits {\it CharCNN} when compared to the original model with a plain softmax layer. 
\end{itemize}

\noindent \textbf{Positive results:}
\begin{itemize}
\item The {\it RE+RW} option puts {\it CharCNN}'s performance close to that of the original version, while reducing the model size by 30--75\%.
\item The {\it RE} and {\it RE+RW} are the best reusing options for  {\it SylConcat}, which make it on par with the original {\it CharCNN} model, despite having 35--75\% fewer parameters.
\item The {\it RE} and {\it RE+RW} configurations benefit {\it MorphSum} making it not only better than its original version but also better than all other models and significantly smaller than the word-level model with reused embeddings.
\end{itemize}

\begin{table*}[h]
\setlength{\tabcolsep}{3pt}
\begin{center}
\footnotesize
\begin{tabular}{l l | c c c c c | c c c}
\cline{2-10}
 & \multirow{2}{*}{Model} & \multicolumn{5}{c|}{In Vocabulary}  & \multicolumn{3}{c}{Out-of-Vocabulary}\\
& & \textit{while} & \textit{his} & \textit{you} & \textit{richard} & \textit{trading} & \textit{computer-aided} & \textit{misinformed} & \textit{looooook} \\
\cline{2-10}

\parbox[t]{0.7mm}{\multirow{12}{*}{\rotatebox[origin=c]{90}{INPUT EMBEDDINGS}}}\ \ \ 
 &  & c\underline{hile} & \underline{h}h\underline{s} & g\underline{o}d & gra\underline{ha}m & traded & computer-guided & informed & look\\
& CharCNN & \underline{wh}o\underline{le} & its & we & \underline{har}old & \underline{tradi}tion & computerized & per\underline{formed} & looks \\
& (original) & meanwhile & her & your & edw\underline{ard} & he\underline{ading} & computer-driven & \underline{formed} & looking \\
& & although & t\underline{his} & i & ronald & ero\underline{ding} & black-and-white & confi\underline{rmed} & looked \\
\cline{2-10}

 &  & although & my & kemp &  thomas & print\underline{ing} & computer-guided & re\underline{infor}ced & ---\\
 & SylConcat & though & \underline{his}toric & welch & robert & work\underline{ing} & computer-driven & surprised & --- \\
 & (original) & when & your & i & stephen & lending & computerized & succeeding & --- \\
 &  & mean & irish & shere & alan & recor\underline{ding} & computer & succeed & --- \\
\cline{2-10}

 &  & although & mystery & i & stephen & program-trading & cross-border & informed & nato \\
 & MorphSum & whenever & my & ghandi & leon\underline{ard} & insider-trading & bank-backed & injured & lesko \\
 & (original) & when & w\underline{his}key & we & william & relations & pro-choice & confined & imo \\
 &  & 1980s & sour & cadillac & robert & insurance & government-owned & \underline{formed} & swapo \\

\cline{2-10}
{\tiny\ }\\
\cline{2-10}

\parbox[t]{0.7mm}{\multirow{4}{*}{\rotatebox[origin=c]{90}{I/O EMB'S}}} &  & t\underline{hi} & her & we & ger\underline{ard} & trades & computer-guided & informed & look  \\
& CharCNN & when & its & your & gerald & trader & large-scale & per\underline{formed} & outlook \\
& + RE + RW & after & the & \underline{you}ng & william & traders & high-quality & outper\underline{formed} & looks \\
&  & above & \underline{h}eir & wh\underline{y} & edw\underline{ard} & trade & futures-related & confi\underline{rmed} & looked \\
\cline{2-10}




\end{tabular}
\end{center}
\vspace{-10pt}\caption{Nearest neighbors based on cosine similarity. We underline character ngrams in words which are close to the given word orthographically rather than semantically. The \texttt{pyphen} syllabifier, which is used in {\it SylConcat}, failed to segment the word `looooook' into syllables, and therefore its neighbors are not available.}
\label{neighbors}
\end{table*}

\noindent In what follows we proceed to analyze the obtained results.

\subsection{CharCNN is biased towards surface form}
We hypothesize that the reason {\it CharCNN} does not benefit from tied weights is that CNN over character embeddings is an excessively flexible model which learns to adapt to a surface form more than to semantics. To validate this hypothesis we pick several words\footnote{We pick the same words as \citeauthor{kim2016character} \shortcite{kim2016character}.} from the English PTB vocabulary and consider their nearest neighbors under cosine similarity as produced by the medium-sized models (with the regular softmax layer) at input (Table \ref{neighbors}). As we can see from the  examples, the {\it CharCNN} model is somewhat more biased towards surface forms at input than {\it SylConcat} and {\it MorphSum}.\footnote{A similar observation for character-aware NLMs was made by \citeauthor{vania2017characters} \shortcite{vania2017characters}.} When {\it CharCNN} is reused to generate a softmax embedding matrix this bias is propagated to output embeddings as well (Table~\ref{neighbors}). 

\subsection{Tying weights bottom-up}
From Table \ref{results} one can notice that tying weights  without tying subword embeddings ({\it RW}) \textit{always} results in worse performance than the tying both weights and embeddings ({\it RE+RW}). Recall that subword embedding lookup is done before the weights of subword-aware embedding model are used (see Figure \ref{architecture}). This leads us to the following 
\begin{conjecture*}
Let $\mathbf{E}^\text{in}_\mathcal{S}=\mathbf{\Theta}^\text{in}_0, \mathbf{\Theta}^\text{in}_1, \mathbf{\Theta}^\text{in}_2, \ldots, \mathbf{\Theta}^\text{in}_n$ be the parameters of the consecutive layers of a subword-aware input embedding model (\ref{emb_model}), i.e. 
$
\mathbf{x} = \mathbf{x}^{(n)} = f_n\left(\mathbf{x}^{(n-1)};\mathbf{\Theta}^\text{in}_n\right)$, \ldots, $\mathbf{x}^{(1)} = f_1\left(\mathbf{x}^{(0)};\mathbf{\Theta}^\text{in}_1\right)$, $\mathbf{x}^{(0)} = f_0\left(\sigma(w);\mathbf{E}^\text{in}_\mathcal{S}\right)$
and let $\mathbf{E}^\text{out}_\mathcal{S}=\mathbf{\Theta}^\text{out}_0, \mathbf{\Theta}^\text{out}_1, \mathbf{\Theta}^\text{out}_2, \ldots, \mathbf{\Theta}^\text{out}_n$ be the parameters of the consecutive layers of a subword-aware embedding model used to generate the output projection matrix (\ref{new_emb}). Let $A$ be a subword-aware neural language model in which the first $(j+1)$ layers of input and output embedding sub-networks have tied weights: 
$$
\forall{i}=\overline{0,j}:\quad\mathbf{\Theta}^\text{in}_i=\mathbf{\Theta}^\text{out}_i ,
$$
and let $B$ be a model in which at least one layer below the $(j+1)^\text{th}$ layer has untied weights:
$$
\exists{i}=\overline{0,j-1}:\quad\mathbf{\Theta}^\text{in}_i\ne\mathbf{\Theta}^\text{out}_i,\ \mathbf{\Theta}^\text{in}_j=\mathbf{\Theta}^\text{out}_j.
$$
Then model $B$ performs at most as well as model $A$, i.e. $\text{PPL}_{A} \le \text{PPL}_{B}$.
\end{conjecture*}
\noindent To test this conjecture empirically, we conduct the following experiments: in all three embedding models ({\it CharCNN}, {\it SylConcat}, and {\it MorphSum}), we reuse different combinations of layers. 
If an embedding model has $n$ layers, there are $2^n$ ways to reuse them, as each layer can either be tied or untied at input and output. However, there are two particular configurations for each of the embedding models that do not interest us: (i) when neither of the layers is reused, or (ii) when only the very first embedding layer is reused. Hence, for each model we need to check $2^n-2$ configurations. For faster experimentation we evaluate only small-sized models on PTB.
\begin{table*}[h]
\begin{small}
\begin{center}
\begin{tabular}{c c c c c l}
\hline
HW$_2$     & HW$_1$     & CNN        & Emb      & PPL &         \\
\hline
           &            & \checkmark &          & \mybar{94.1}{85}{4.0}\\
           &            & \checkmark &\checkmark& \bestbar{92.8}{85}{4.0}   \\
\hline
           & \checkmark &            &          & \mybar{94.6}{85}{4.0}\\
           & \checkmark &            &\checkmark& \mybar{94.5}{85}{4.0}\\
           & \checkmark & \checkmark &          & \mybar{93.1}{85}{4.0}\\ 
           & \checkmark & \checkmark &\checkmark& \bestbar{90.1}{85}{4.0}\\ 
\hline
\checkmark &            &            &          & \mybar{94.9}{85}{4.0}\\ 
\checkmark &            &            &\checkmark& \mybar{99.2}{85}{4.0}\\ 
\checkmark &            & \checkmark &          & \mybar{94.1}{85}{4.0}\\ 
\checkmark &            & \checkmark &\checkmark& \mybar{92.5}{85}{4.0}\\ 
\checkmark & \checkmark &            &          & \mybar{94.3}{85}{4.0}\\ 
\checkmark & \checkmark &            &\checkmark& \mybar{97.8}{85}{4.0}\\ 
\checkmark & \checkmark & \checkmark &          & \mybar{96.3}{85}{4.0}\\ 
\checkmark & \checkmark & \checkmark &\checkmark& \bestbar{91.0}{85}{4.0}\\ 
\hline
\end{tabular}\hspace{20pt}
\begin{tabular}{c c c c l}
\hline
HW$_2$     & HW$_1$     & Emb      & PPL &         \\
\hline
           & \checkmark &          & \mybar{95.4}{80}{2.5}  \\
           & \checkmark &\checkmark& \bestbar{87.4}{80}{2.5}\\  
\hline
\checkmark &            &          & \mybar{99.0}{80}{2.5}  \\
\checkmark &            &\checkmark& \bestbar{87.9}{80}{2.5}  \\
\checkmark & \checkmark &          & \mybar{96.2}{80}{2.5}  \\
\checkmark & \checkmark &\checkmark& \mybar{88.4}{80}{2.5}\\
\hline
 & & & & \\
\hline
HW$_2$     & HW$_1$     & Emb      & PPL &                  \\
\hline
           & \checkmark &          & \mybar{90.0}{80}{2}     \\
           & \checkmark &\checkmark& \bestbar{84.7}{80}{2}\\           
\hline
\checkmark &            &          & \mybar{89.9}{80}{2} \\
\checkmark &            &\checkmark& \mybar{85.7}{80}{2}  \\
\checkmark & \checkmark &          & \mybar{89.4}{80}{2} \\
\checkmark & \checkmark &\checkmark& \bestbar{85.1}{80}{2}\\
\hline
\end{tabular}
\end{center}
\end{small}
\vspace{-10pt}
\caption{Reusing different combinations of layers in small {\it CharCNN} (left), small {\it SylConcat} (top right) and small {\it MorphSum} on PTB data. ``\checkmark'' means that the layer is reused at output.}
\label{morph_sum_reuse}
\end{table*}
\begin{figure*}[h]
\begin{center}
\includegraphics[scale=0.42]{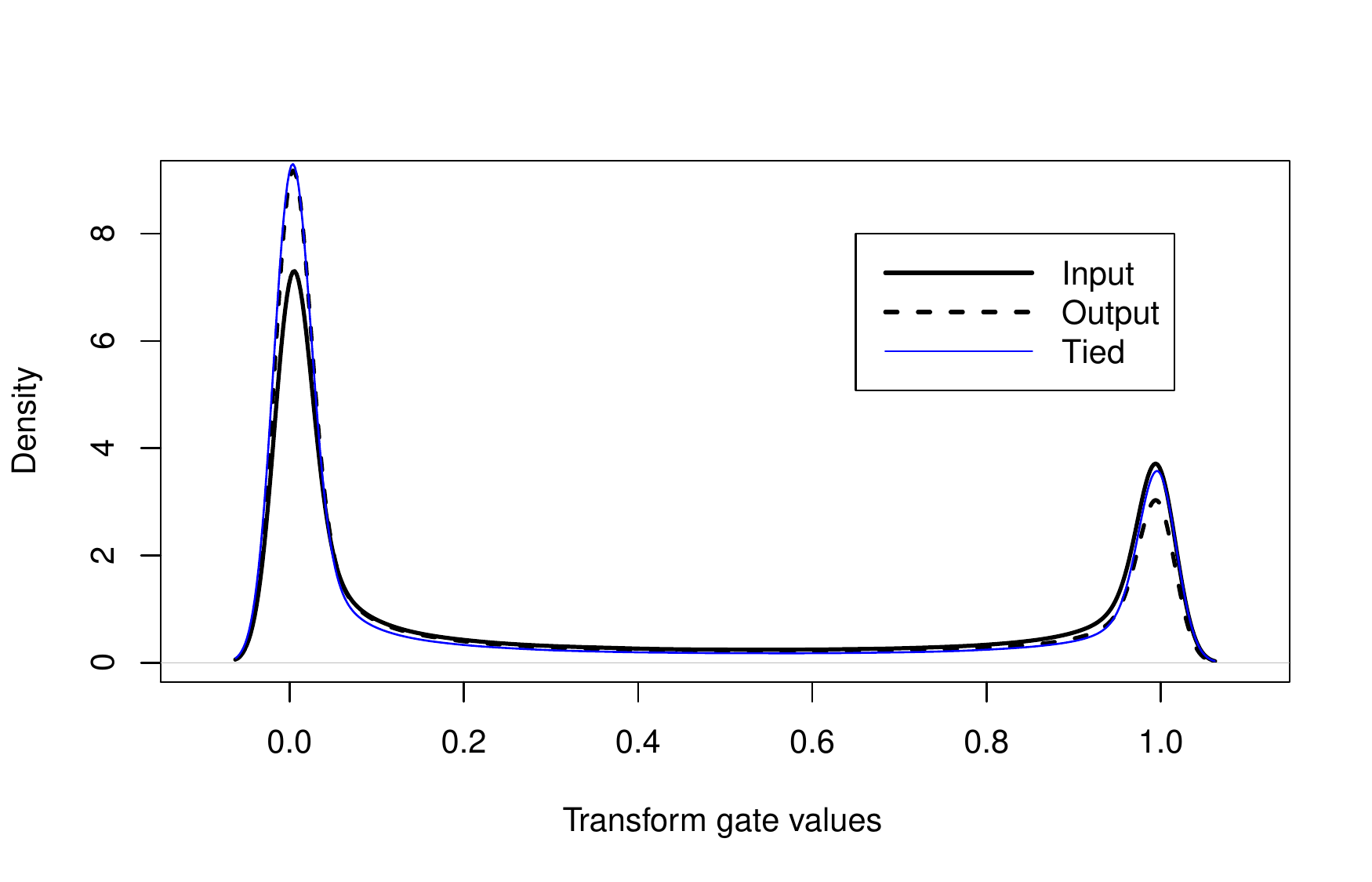}\qquad
\includegraphics[scale=0.42]{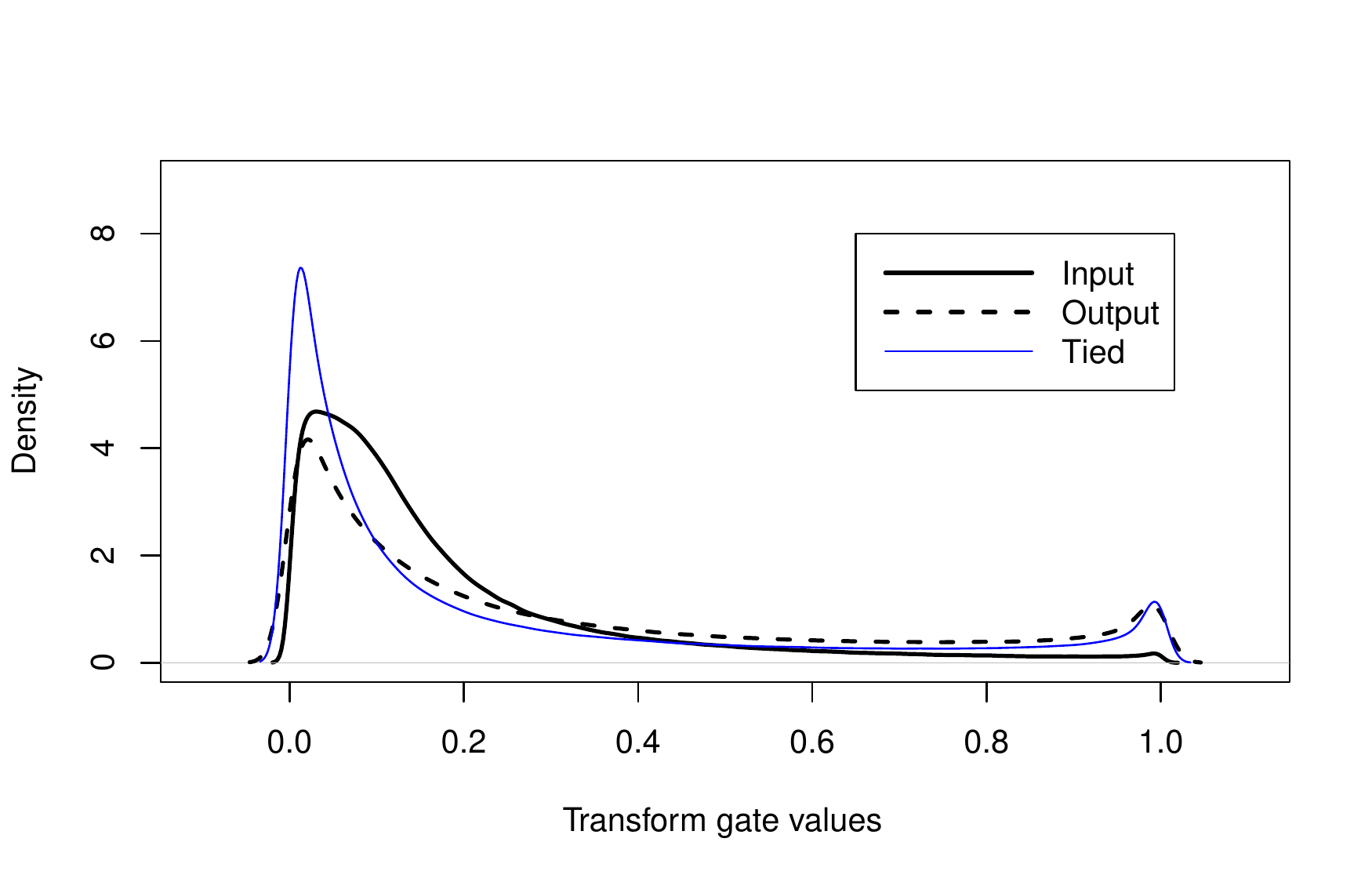}
\end{center}
\vspace{-10pt}\caption{Kernel density estimations of the transform gate values of the first  highway layers in {\it SylConcat} (left) and {\it MorphSum}. Values corresponding to `Input' and `Output' curves come from the HW$_2$+Emb configurations, while those corresponding to `Tied' curves come from the HW$_2$+HW$_1$+Emb configurations.}
\label{t_gates}
\end{figure*}
\begin{figure*}[h]
\includegraphics[scale=0.22]{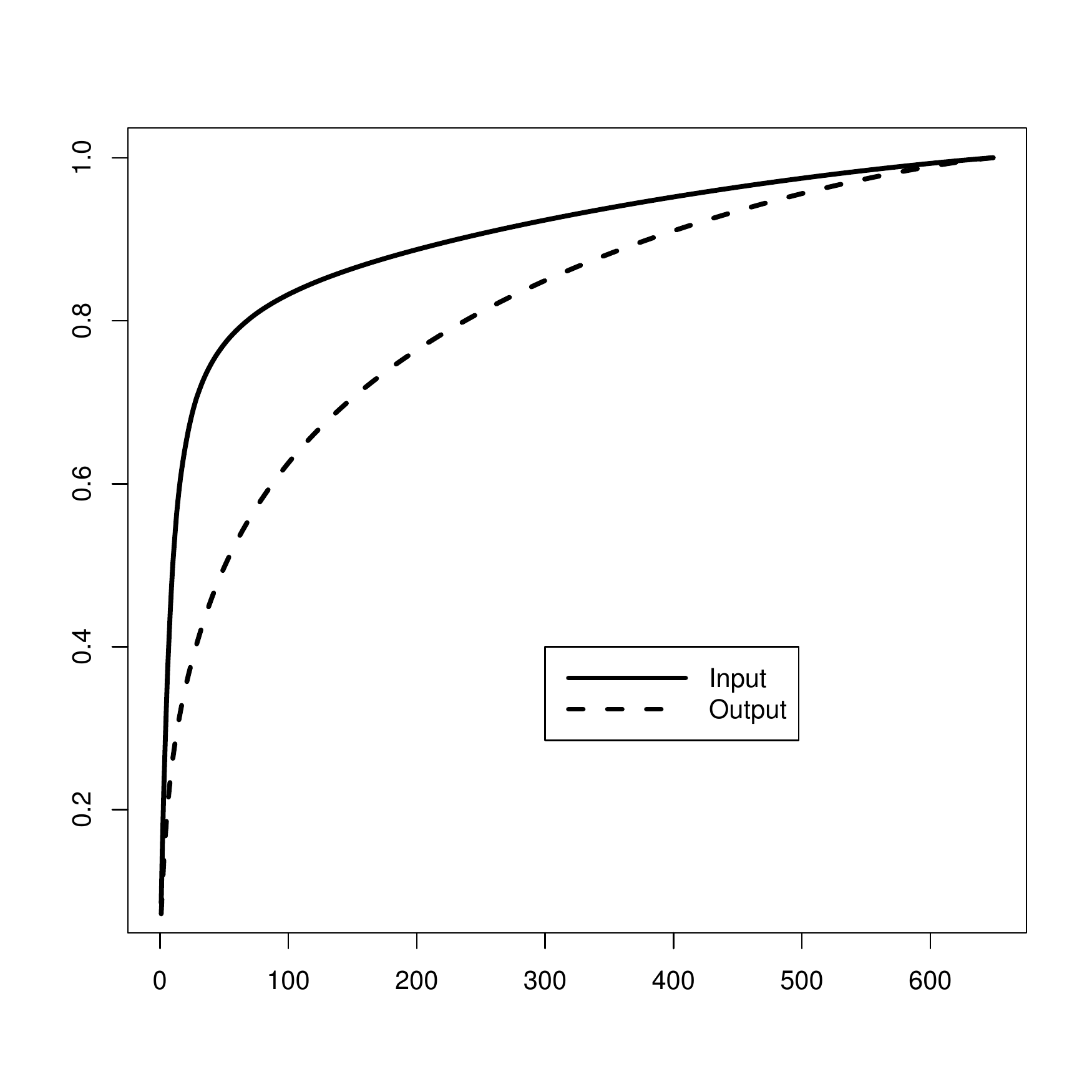} \includegraphics[scale=0.22]{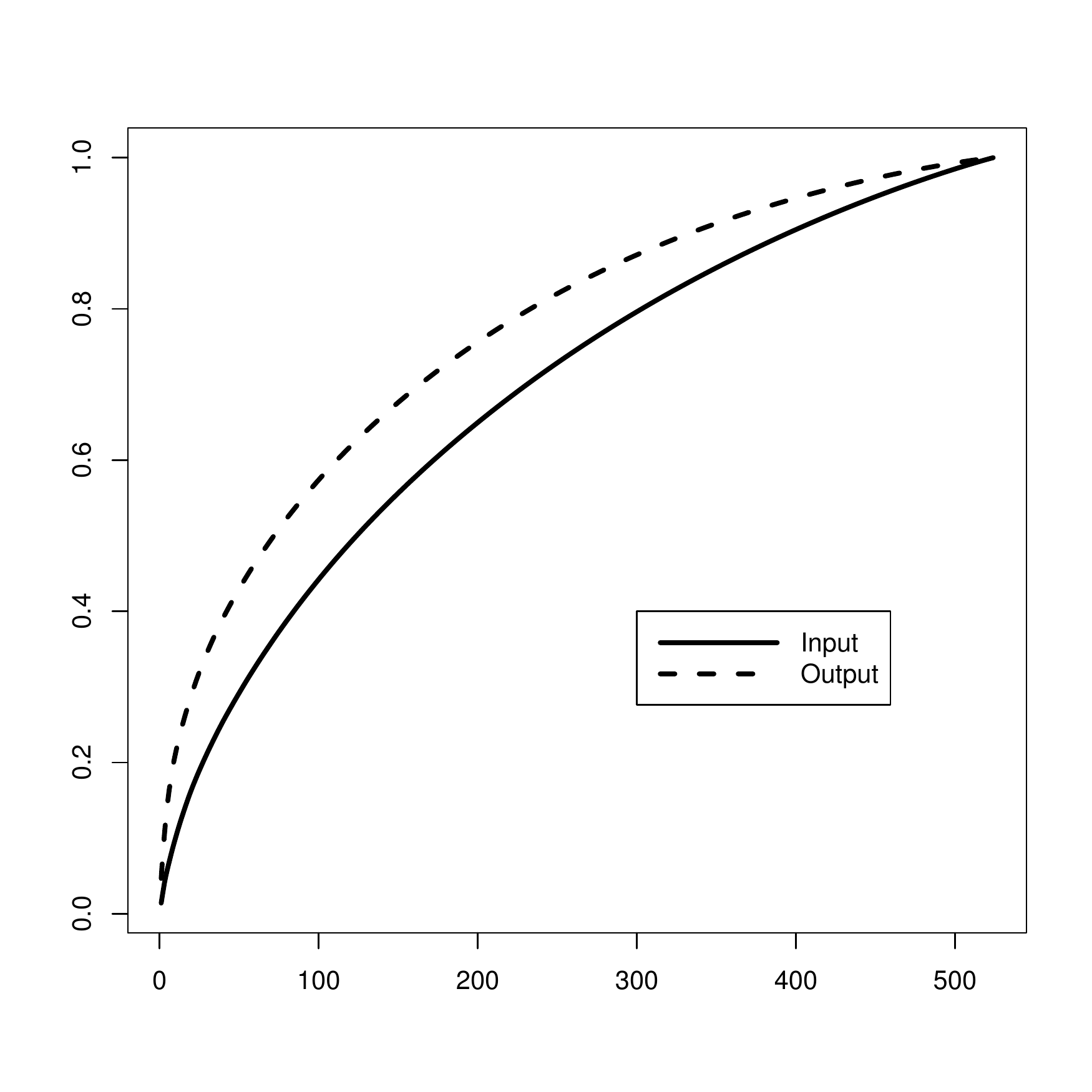} \includegraphics[scale=0.22]{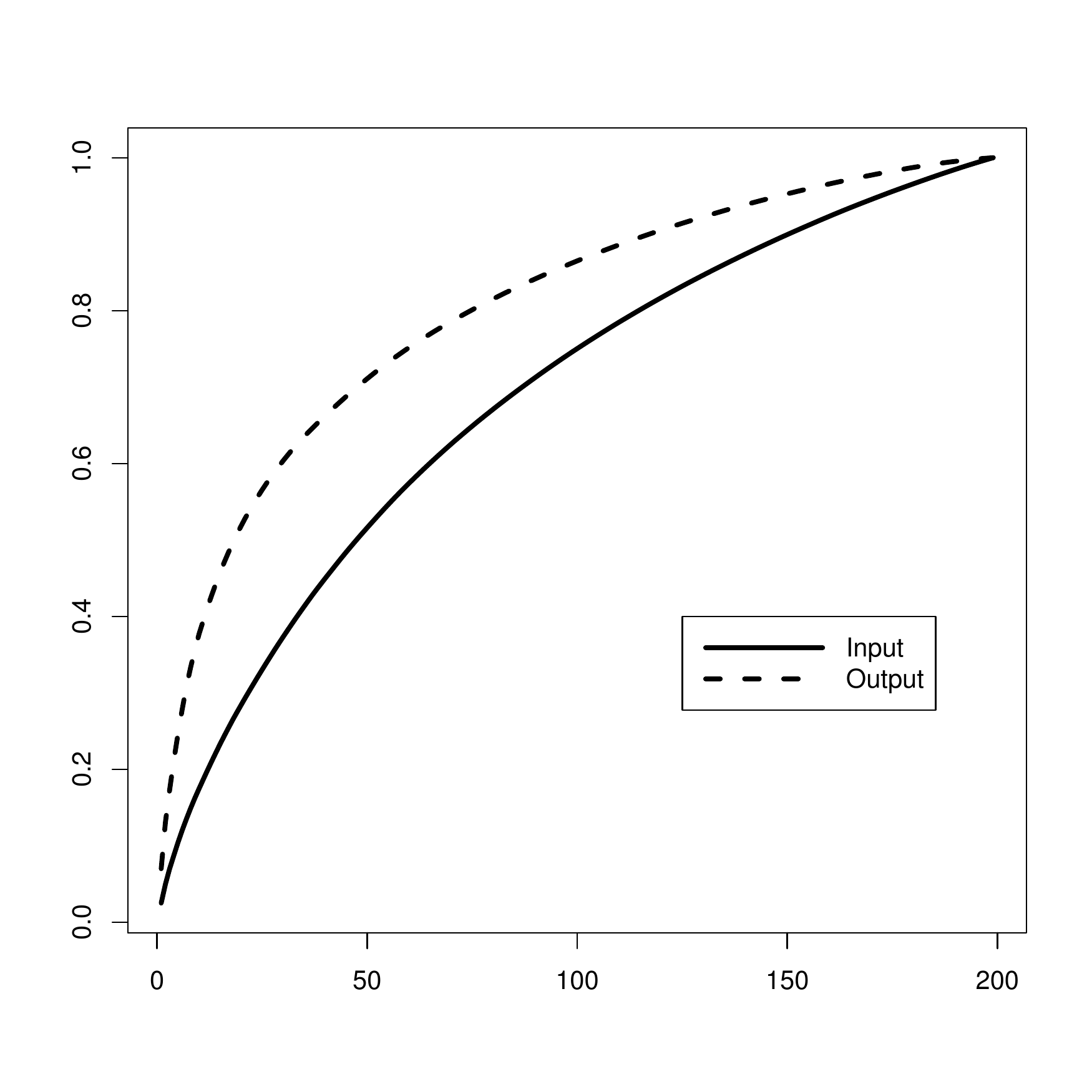} \includegraphics[scale=0.22]{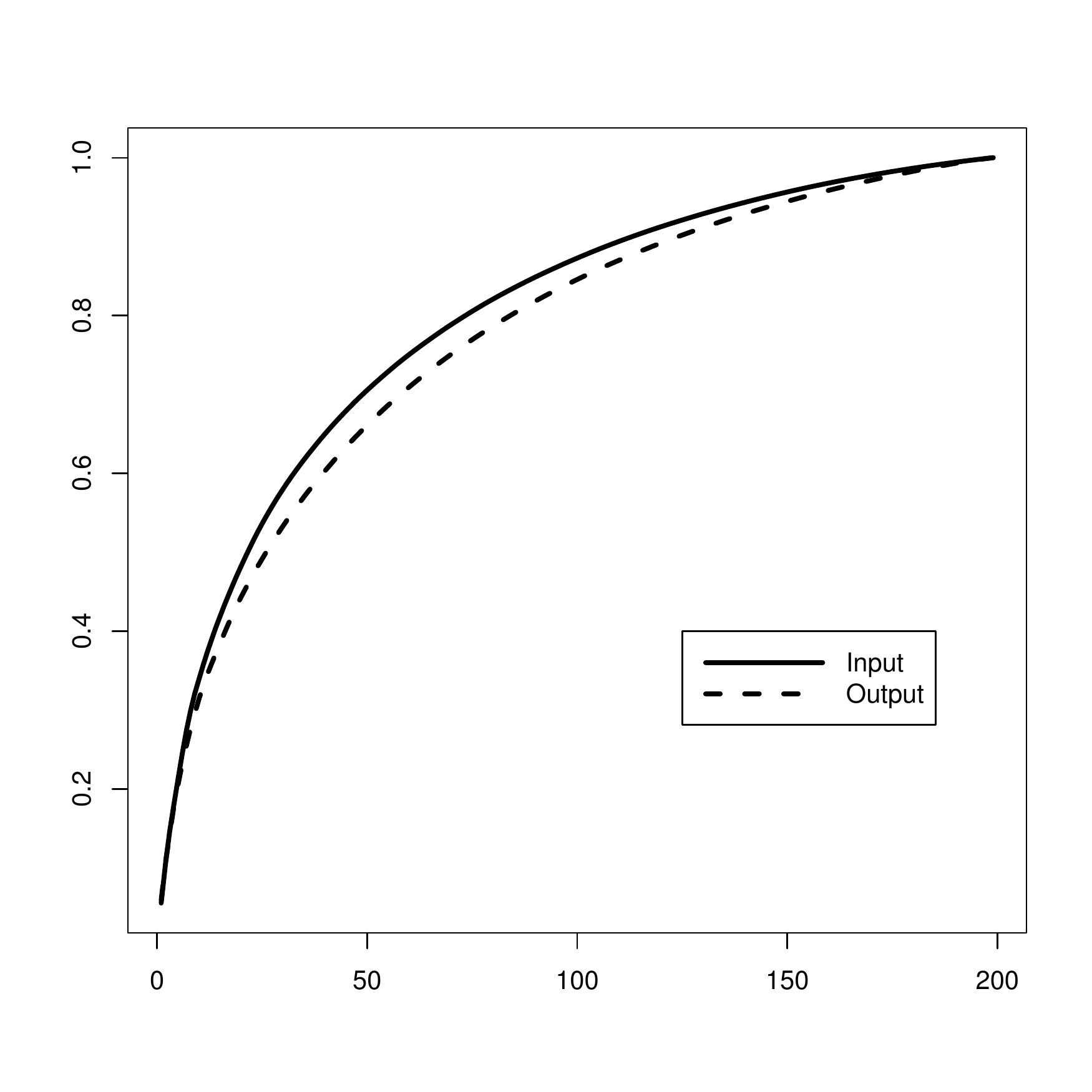}
\vspace{-10pt}\caption{PCA applied to input and word embeddings produced by different models. Horizontal axis corresponds to number of principal components, vertical axis corresponds to percentage of total variance to retain. From left to right: word-level model, {\it CharCNN}, {\it SylConcat}, {\it MorphSum}.}
\label{pca}
\end{figure*}
The results are reported in Table  \ref{morph_sum_reuse}. As we can see, the experiments in general reject our conjecture: in {\it SylConcat} leaving an untied first highway layer between tied embedding and second highway layers (denote this as {HW$_2$+Emb}) turned out to be slightly better than tying all three layers ({HW$_2$+HW$_1$+Emb}). Recall, that a highway is a weighted average between nonlinear and identity transformations of the incoming vector:
$$
\mathbf{x} \mapsto \mathbf{t}\odot\text{ReLU}(\mathbf{x}\mathbf{A}+\mathbf{b})+(\mathbf{1}-\mathbf{t})\odot\mathbf{x},
$$ 
where $\mathbf{t}=\sigma(\mathbf{x}\mathbf{W}+\mathbf{c})$ is a transform gate, $\mathbf{A}$, $\mathbf{W}$, $\mathbf{b}$ and $\mathbf{c}$ are trainable parameters, and $\odot$ is the element-wise multiplication operator. To find out why leaving an untied highway below a tied one is beneficial in {\it SylConcat}, we compare the distributions of the transform gate values $\mathbf{t}$ from the first highway layers of both configurations, {HW$_2$+Emb} and HW$_2$+HW$_1$+Emb, in {\it SylConcat} and {\it MorphSum} (Figure \ref{t_gates}). 

We can see that {\it SylConcat}  heavily relies on nonlinearity in the first highway layer, while {\it MorphSum} does not utilize much of it. This means that in MorphSum, the highway is close to an identity operator ($\mathbf{t}\approx\mathbf{0}$), and does not transform the sum of morpheme vectors much, either at input or at output. Therefore, tying the first highway layer is natural to {\it Morh-Sum}. {\it SylConcat}, on the other hand, applies non-linear transformations to the concatenation of syllable vectors, and hence makes additional preparations of the word vector for the needs of the RNNLM at input and for Softmax prediction at output. These needs differ from each other (as shown in the next subsection). This is why {\it SylConcat} benefits from an additional degree of freedom when the first highway is left untied. 

Despite not being true in all cases, and due to being true in many cases, we believe that the above-mentioned conjecture is still useful. In short it can be summarized as a practical hands-on rule: $$\textit{Layers should be tied consecutively bottom-up},$$ i.e. one should not leave untied layer(s) below a tied one. Keep in mind that this rule does not guarantee a performance increase as more and more layers are tied. It only says that leaving untied weights below the tied ones is likely to be worse than not doing so. 

\subsection{Difference between input and output embeddings}
One can notice from the results of our experiments (Table \ref{morph_sum_reuse}) that having an untied second highway layer above the first one always leads to better performance than when it is tied. This means that there is a benefit in letting  word embeddings slightly differ at input and output, i.e. by specializing them for the needs of RNNLM at input and of Softmax at output. This specialization is quite natural, as input and output representations of words have two different purposes: input representations send a signal to the RNNLM about the current word in a sequence, while output representations are needed to predict the next word given \textit{all} the preceding words. 
The difference between input and output word representations is discussed in greater detail by \citet{garten2015combining} and \citet{press2016using}. Here we decided to verify the difference  indirectly: we test whether intrinsic dimensionality of word embeddings significantly differs at input and output. For this, we apply principal component analysis to word embeddings produced by all models in ``no reusing'' mode. The results are given in Figure \ref{pca}, where we can see that dimensionalities of input and output embeddings differ in the word-level model, {\it CharCNN}, and {\it SylConcat} models, but the difference is less significant in {\it MorphSum} model. Interestingly, in word-level and {\it MorphSum} models the output embeddings have more principal components than the input ones. In {\it CharCNN} and {\it SylConcat}, however, results are to other  way around. We defer the study of this phenomenon to the future work.

\subsection{CharCNN generalizes better than MorphSum}
One may expect larger units to work better than smaller units, but smaller units to generalize better than larger units. This certainly depends on how one defines generalizability of a language model. If it is an ability to model unseen text with \textit{unseen} words, then, indeed, character-aware models may perform better than syllable- or morpheme-aware ones. This can be partially seen from Table~\ref{neighbors}, where the OOV words are better handled by \textit{CharCNN} in terms of in-vocabulary nearest neighbors. However, to fully validate the above-mentioned expectation we conduct additional experiments: we train two models, \textit{CharCNN} and \textit{MorphSum}, on PTB and then we evaluate them on the test set of Wikitext-2 (245K words, 10K word-types). Some words in Wikitext-2 contain characters or morphemes that are not present in PTB, and therefore such words cannot be embedded by \textit{CharCNN} or \textit{MorphSum} correspondingly. Such words were replaced by the \texttt{<unk>} token, and we call them new OOVs\footnote{These are ``new'' OOVs, since the original test set of Wikitext-2 already contains ``old'' OOVs marked as \texttt{<unk>}.}. The results of our experiments are reported in Table~\ref{generalizability}.
\begin{table}[h]
\begin{tabular}{l c c}
\toprule
Model & \# new OOVs & PPL \\
\midrule
CharCNN + RE + RW & 3659 & 306.8 \\
MorphSum + RE + RW & 4195 & 316.2\\
\bottomrule
\end{tabular}
\caption{Training on PTB and testing on Wikitext-2.}
\label{generalizability}
\end{table}
Indeed, \textit{CharCNN} faces less OOVs on unseen text, and thus generalizes better than \textit{MorphSum}. 

\subsection{Performance on non-English Data}
\begin{figure}[h]
\begin{center}
\includegraphics[width=0.4\textwidth]{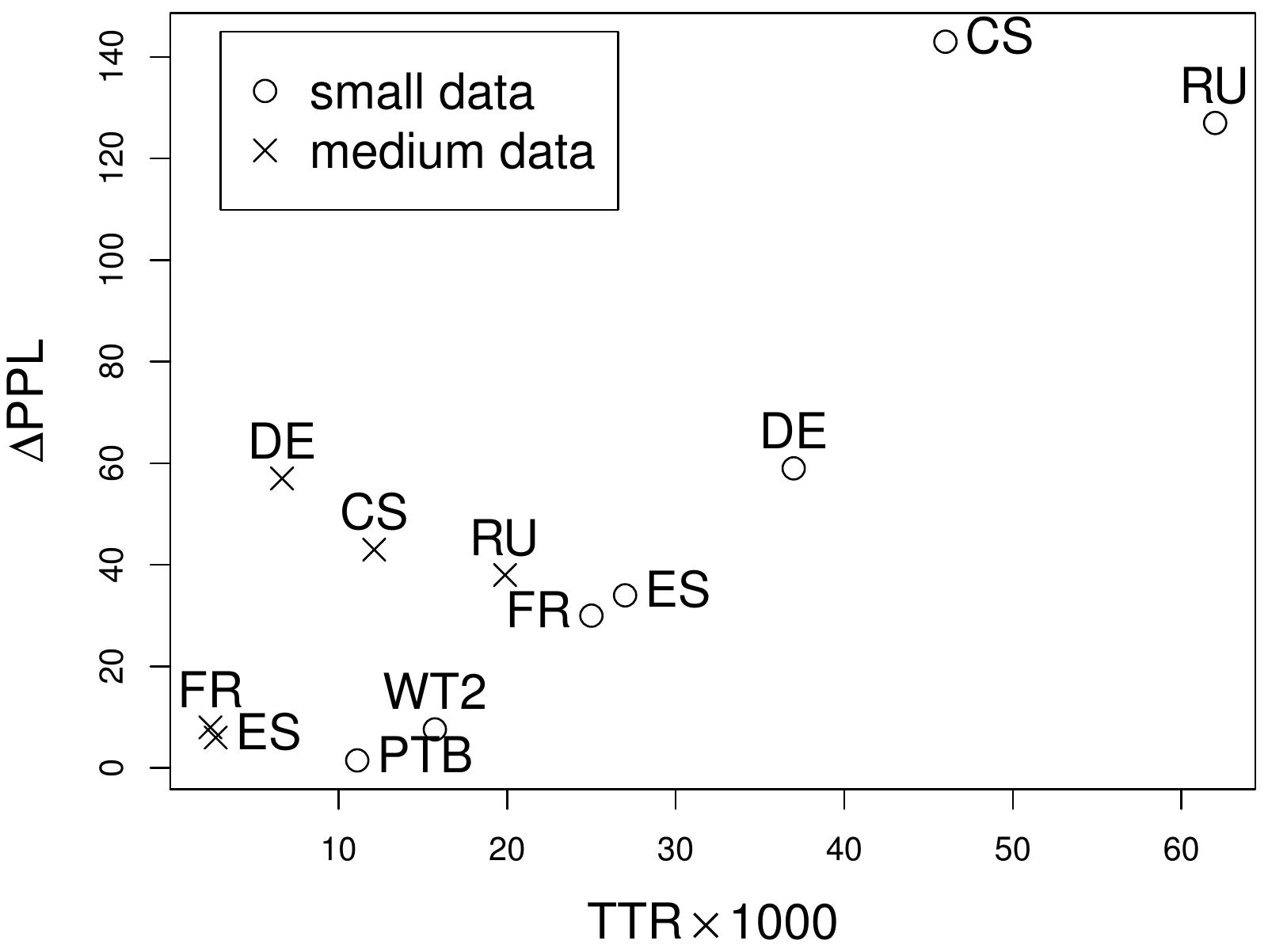}
\end{center}
\vspace{-10pt}\caption{PPL improvement vs TTR. $\Delta\text{PPL}=\text{PPL}_\text{Word+RE}-\text{PPL}_\text{MorphSum+RE+RW}$.}
\label{ppl_ttr}
\end{figure}

According to Table \ref{results}, {\it MorphSum+RE+RW} comfortably outperforms the strong baseline {\it Word+RE} \cite{inan2016tying}. It is interesting to see whether this advantage extends to non-English languages which have richer morphology. For this purpose we conduct evaluation of both  models on small (1M tokens) and medium (17M--51M tokens) data in five languages (see corpora statistics in Appendix~\ref{sizes}). Due to hardware constraints we only train the small
models on medium-sized data. We used the same architectures for all languages and did not perform any language-specific tuning of hyperparameters, which are specified in Appendix \ref{opt}. The results are provided in Table \ref{evaluation}.
\begin{table}[t]
\setlength{\tabcolsep}{4pt}
\begin{center}
\begin{small}
\begin{tabular}{c l c c c c c c}
\toprule
& {Model} & FR & ES & DE & CS & RU\\
\midrule
\parbox[t]{0.5mm}{\multirow{2}{*}{\rotatebox[origin=c]{90}{S}}} & Word+RE  & 218 & 205 & 305 & 514 & 364 & \parbox[t]{0.5mm}{\multirow{4}{*}{\rotatebox[origin=c]{90}{D-S}}} \\
& MorphSum+RE+RW & \textbf{188} & \textbf{171} & \textbf{246} & \textbf{371} & \textbf{237} & \\
\cline{1-7}
\parbox[t]{0.5mm}{\multirow{2}{*}{\rotatebox[origin=c]{90}{M}}} & Word+RE  & 205 & 193 & 277 & 488 & 351 & \\
& MorphSum+RE+RW & \textbf{172} & \textbf{157} & \textbf{222} & \textbf{338} & \textbf{210} & \\
\midrule
\parbox[t]{0.5mm}{\multirow{2}{*}{\rotatebox[origin=c]{90}{S}}} & Word+RE & 167 & 149 & 285 & 520 & 267 & \parbox[t]{0.5mm}{\multirow{2}{*}{\rotatebox[origin=c]{90}{D-M}}} \\
& MorphSum+RE+RW & \textbf{159} & \textbf{143} & \textbf{242} & \textbf{463} & \textbf{229} & \\
\bottomrule
\end{tabular}
\end{small}
\end{center}
\vspace{-10pt}\caption{Evaluation on non-English data. {\it MorphSum+RE+RW} has significantly less parameters than {\it Word+RE} (Appendix \ref{sizes}). S --- small model, M --- medium model, D-S --- small data, D-M --- medium data; FR --- French, ES --- Spanish, DE --- German, CS --- Czech, RU --- Russian.}\vspace{-10pt}
\label{evaluation}
\end{table}
\begin{table*}[h]
\begin{center}
\begin{tabular}{l c c c c c c c}
\toprule
Model & PTB & WT-2 & CS & DE & ES & FR & RU \\
\midrule
AWD-LSTM-Word w/o emb. dropout & 61.38 & 68.50 & 410 & 241 & 145 & 151 & 232 \\
AWD-LSTM-MorphSum + RE + RW & 61.17 & 66 .92 & 253 & 177 & 126 & 140 & 162 \\
\bottomrule
\end{tabular}
\end{center}
\caption{Replacing LSTM with AWD-LSTM.}
\label{awd_lstm}
\end{table*}
As one can see, the advantage of the morpheme-aware model over the word-level one is even more pronounced for non-English data. Also, we can notice that the gain is larger for small data sets. We hypothesize that the advantage of {\it MorphSum+RE+RW} over {\it Word+RE} diminishes with the decrease of type-token ratio (TTR). A scatterplot of PPL change versus TTR (Figure~\ref{ppl_ttr}) supports this hypothesis.
Moreover, there is a strong correlation between these two quantities: $\hat{\rho}(\Delta{\rm PPL}, {\rm TTR})=0.84$, i.e. one can predict the mean decrease in PPL from the TTR of a text with a simple linear regression: $$\Delta{\rm PPL}\approx2,109\times{\rm TTR}.$$

\subsection{Replacing LSTM with AWD-LSTM}
The empirical perplexities in Table~\ref{results} are way above the current state-of-the-art on the same datasets \cite{melis2017state}. However, the  approach of \newcite{melis2017state} requires thousands of evaluations and is feasible for researchers who have access to hundreds of GPUs. Unfortunately, we do not have such access. Also, the authors do not disclose the optimal hyperparameters they found, and thus we could not reproduce their models. There is another state-of-the-art language model, AWD-LSTM \cite{merity2017regularizing}, which has open-source code. We replaced this model's word embedding layer with the \textit{MorphSum} subnetwork and fully reused morpheme embeddings and other weights of \textit{MorphSum} at output. We refer to such modification as \textit{AWD-LSTM-MorphSum + RE + RW}. We trained both models without fine-tuning (due to time constraints) and we did not use embedding dropout (section 4.3 of \newcite{merity2017regularizing}) in either model, as it is not obvious how embeddings should be dropped in the case of \textit{AWD-LSTM-MorphSum}. The results of evaluation on the PTB, Wikitext-2, and non-English datasets are given in Table~\ref{awd_lstm}. 

Although \textit{AWD-LSTM-MorphSum} is on par with \textit{AWD-LSTM-Word} on PTB and is slightly better on Wikitext-2, replacing plain word embeddings with the subword-aware model with appropriately reused parameters is crucial for non-English data. Notice that AWD-LSTM underperforms LSTM (used by us) on Czech dataset  (cf. Table~\ref{evaluation}). We think that the hyperparameters of AWD-LSTM in \newcite{merity2017regularizing} are thoroughly tuned for PTB and Wikitext-2 and may poorly generalize to other datasets.

\section{Conclusion}
There is no single best way to reuse parameters in all subword-aware neural language models: the reusing method should be tailored to each type of subword unit and embedding model. However, instead of testing an exponential (w.r.t. sub-network depth) number of configurations, it is sufficient to check only those where weights are tied consecutively bottom-up. 

Despite being similar, input and output embeddings solve different tasks. Thus, fully tying input and output embedding sub-networks in subword-aware neural language models is worse than letting them be slightly different. This raises the question whether the same is true for pure word-level models, and we defer its study to our future work.

One of our best configurations, a simple morpheme-aware model which sums morpheme embeddings and fully reuses the embedding sub-network, outperforms the competitive word-level language model while significantly reducing the number of trainable parameters. However, the performance gain diminishes with the increase of training set size.

\section*{Acknowledgements}
We gratefully acknowledge the NVIDIA Corporation for their donation of the Titan X Pascal GPU used for this research. The work of Zhenisbek Assylbekov has been funded by the Committee of Science of the Ministry of Education and Science of the Republic of Kazakhstan, contract \# 346/018-2018/33-28, IRN AP05133700. The authors would like to thank anonymous reviewers for their valuable feedback, and Dr. J.~N.~Washington for proofreading an early version of the paper.

\bibliography{naaclhlt2018}

\begin{thebibliography}{}
\expandafter\ifx\csname natexlab\endcsname\relax\def\natexlab#1{#1}\fi

\bibitem[{Assylbekov et~al.(2017)Assylbekov, Takhanov, Myrzakhmetov, and
  Washington}]{assylbekov2017syllable}
Zhenisbek Assylbekov, Rustem Takhanov, Bagdat Myrzakhmetov, and Jonathan~N.
  Washington. 2017.
\newblock Syllable-aware neural language models: A failure to beat
  character-aware ones.
\newblock In {\em Proc. of EMNLP\/}.

\bibitem[{Bengio et~al.(2001)Bengio, Ducharme, Vincent, and
  Jauvin}]{bengio2001neural}
Yoshua Bengio, R{\'e}jean Ducharme, Pascal Vincent, and Christian Jauvin. 2001.
\newblock \href{http://www.iro.umontreal.ca/˜lisa/pointeurs/nips00_lm.ps}{A
  neural probabilistic language model}
  \url{http://www.iro.umontreal.ca/˜lisa/pointeurs/nips00_lm.ps}.

\bibitem[{Botha and Blunsom(2014)}]{botha2014compositional}
Jan Botha and Phil Blunsom. 2014.
\newblock Compositional morphology for word representations and language
  modelling.
\newblock In {\em Proc. of ICML\/}.

\bibitem[{Chen et~al.(2016)Chen, Grangier, and Auli}]{DBLP:conf/acl/ChenGA16}
Wenlin Chen, David Grangier, and Michael Auli. 2016.
\newblock Strategies for training large vocabulary neural language models.
\newblock In {\em Proc. of ACL\/}.

\bibitem[{Cotterell and Sch{\"u}tze(2015)}]{cotterell2015morphological}
Ryan Cotterell and Hinrich Sch{\"u}tze. 2015.
\newblock Morphological word-embeddings.
\newblock In {\em Proc. of HLT-NAACL\/}.

\bibitem[{Gal and Ghahramani(2016)}]{gal2016theoretically}
Yarin Gal and Zoubin Ghahramani. 2016.
\newblock A theoretically grounded application of dropout in recurrent neural
  networks.
\newblock In {\em Proc. of NIPS\/}.

\bibitem[{Garten et~al.(2015)Garten, Sagae, Ustun, and
  Dehghani}]{garten2015combining}
Justin Garten, Kenji Sagae, Volkan Ustun, and Morteza Dehghani. 2015.
\newblock Combining distributed vector representations for words.
\newblock In {\em In Proc. of VS@HLT-NAACL\/}.

\bibitem[{Hochreiter and Schmidhuber(1997)}]{hochreiter1997long}
Sepp Hochreiter and J{\"u}rgen Schmidhuber. 1997.
\newblock Long short-term memory.
\newblock {\em Neural computation\/} 9(8):1735--1780.

\bibitem[{Inan et~al.(2017)Inan, Khosravi, and Socher}]{inan2016tying}
Hakan Inan, Khashayar Khosravi, and Richard Socher. 2017.
\newblock Tying word vectors and word classifiers: A loss framework for
  language modeling.
\newblock In {\em Proc. of ICLR\/}.

\bibitem[{Jean et~al.(2015)Jean, Cho, Memisevic, and
  Bengio}]{DBLP:journals/corr/JeanCMB14}
S{\'{e}}bastien Jean, Kyunghyun Cho, Roland Memisevic, and Yoshua Bengio. 2015.
\newblock On using very large target vocabulary for neural machine translation.
\newblock In {\em Proc. of ACL-IJCNLP\/}.

\bibitem[{Jozefowicz et~al.(2016)Jozefowicz, Vinyals, Schuster, Shazeer, and
  Wu}]{jozefowicz2016exploring}
Rafal Jozefowicz, Oriol Vinyals, Mike Schuster, Noam Shazeer, and Yonghui Wu.
  2016.
\newblock Exploring the limits of language modeling.
\newblock {\em arXiv preprint arXiv:1602.02410\/} .

\bibitem[{Jozefowicz et~al.(2015)Jozefowicz, Zaremba, and
  Sutskever}]{jozefowicz2015empirical}
Rafal Jozefowicz, Wojciech Zaremba, and Ilya Sutskever. 2015.
\newblock An empirical exploration of recurrent network architectures.
\newblock In {\em Proc. of ICML\/}.

\bibitem[{Kim et~al.(2016)Kim, Jernite, Sontag, and Rush}]{kim2016character}
Yoon Kim, Yacine Jernite, David Sontag, and Alexander~M Rush. 2016.
\newblock Character-aware neural language models.
\newblock In {\em Proc. of AAAI\/}.

\bibitem[{Labeau and Allauzen(2017)}]{labeau2017character}
Matthieu Labeau and Alexandre Allauzen. 2017.
\newblock Character and subword-based word representation for neural language
  modeling prediction.
\newblock In {\em Proc. of SCLeM@EMNLP\/}.

\bibitem[{Liang(1983)}]{liang1983word}
Franklin~Mark Liang. 1983.
\newblock {\em Word Hy-phen-a-tion by Com-put-er\/}.
\newblock Citeseer.

\bibitem[{Ling et~al.(2015)Ling, Dyer, Black, Trancoso, Fermandez, Amir,
  Marujo, and Luis}]{ling-EtAl:2015:EMNLP2}
Wang Ling, Chris Dyer, Alan~W Black, Isabel Trancoso, Ramon Fermandez, Silvio
  Amir, Luis Marujo, and Tiago Luis. 2015.
\newblock Finding function in form: Compositional character models for open
  vocabulary word representation.
\newblock In {\em Proc. of EMNLP\/}.

\bibitem[{Marcus et~al.(1993)Marcus, Marcinkiewicz, and
  Santorini}]{marcus1993building}
Mitchell~P Marcus, Mary~Ann Marcinkiewicz, and Beatrice Santorini. 1993.
\newblock Building a large annotated corpus of english: The penn treebank.
\newblock {\em Computational linguistics\/} 19(2):313--330.

\bibitem[{Melis et~al.(2018)Melis, Dyer, and Blunsom}]{melis2017state}
G{\'a}bor Melis, Chris Dyer, and Phil Blunsom. 2018.
\newblock On the state of the art of evaluation in neural language models.
\newblock In {\em Proc. of ICLR\/}.

\bibitem[{Merity et~al.(2018)Merity, Keskar, and
  Socher}]{merity2017regularizing}
Stephen Merity, Nitish~Shirish Keskar, and Richard Socher. 2018.
\newblock Regularizing and optimizing lstm language models .

\bibitem[{Merity et~al.(2017)Merity, Xiong, Bradbury, and
  Socher}]{merity2016pointer}
Stephen Merity, Caiming Xiong, James Bradbury, and Richard Socher. 2017.
\newblock Pointer sentinel mixture models.
\newblock In {\em Proc. of ICLR\/}.

\bibitem[{Mikolov et~al.(2010)Mikolov, Karafi{\'a}t, Burget, Cernock{\`y}, and
  Khudanpur}]{mikolov2010recurrent}
Tomas Mikolov, Martin Karafi{\'a}t, Lukas Burget, Jan Cernock{\`y}, and Sanjeev
  Khudanpur. 2010.
\newblock Recurrent neural network based language model.
\newblock In {\em Proc. of INTERSPEECH\/}.

\bibitem[{Mikolov et~al.(2012)Mikolov, Sutskever, Deoras, Le, Kombrink, and
  Cernocky}]{mikolov2012subword}
Tom{\'a}{\v{s}} Mikolov, Ilya Sutskever, Anoop Deoras, Hai-Son Le, Stefan
  Kombrink, and Jan Cernocky. 2012.
\newblock Subword language modeling with neural networks.
\newblock {\em preprint (http://www. fit. vutbr. cz/imikolov/rnnlm/char.
  pdf)\/} .

\bibitem[{Mnih and Hinton(2007)}]{mnih2007three}
Andriy Mnih and Geoffrey Hinton. 2007.
\newblock Three new graphical models for statistical language modelling.
\newblock In {\em Proc. of ICML\/}.

\bibitem[{Press and Wolf(2017)}]{press2016using}
Ofir Press and Lior Wolf. 2017.
\newblock Using the output embedding to improve language models.
\newblock In {\em Proc. of EACL\/}.

\bibitem[{Qiu et~al.(2014)Qiu, Cui, Bian, Gao, and Liu}]{qiu2014co}
Siyu Qiu, Qing Cui, Jiang Bian, Bin Gao, and Tie-Yan Liu. 2014.
\newblock Co-learning of word representations and morpheme representations.
\newblock In {\em Proc. of COLING\/}.

\bibitem[{Srivastava et~al.(2015)Srivastava, Greff, and
  Schmidhuber}]{srivastava2015training}
Rupesh~K Srivastava, Klaus Greff, and J{\"u}rgen Schmidhuber. 2015.
\newblock Training very deep networks.
\newblock In {\em Proc. of NIPS\/}.

\bibitem[{Vania and Lopez(2017)}]{vania2017characters}
Clara Vania and Adam Lopez. 2017.
\newblock From characters to words to in between: Do we capture morphology?
\newblock In {\em Proc. of ACL\/}.

\bibitem[{Verwimp et~al.(2017)Verwimp, Pelemans, Wambacq
  et~al.}]{verwimp2017character}
Lyan Verwimp, Joris Pelemans, Patrick Wambacq, et~al. 2017.
\newblock Character-word lstm language models.
\newblock In {\em Proc. of EACL\/}.

\bibitem[{Virpioja et~al.(2013)Virpioja, Smit, Gr{\"o}nroos, Kurimo
  et~al.}]{virpioja2013morfessor}
Sami Virpioja, Peter Smit, Stig-Arne Gr{\"o}nroos, Mikko Kurimo, et~al. 2013.
\newblock Morfessor 2.0: Python implementation and extensions for morfessor
  baseline .

\bibitem[{Werbos(1990)}]{werbos1990backpropagation}
Paul~J Werbos. 1990.
\newblock Backpropagation through time: what it does and how to do it.
\newblock {\em Proc. of the IEEE\/} 78(10):1550--1560.

\bibitem[{Yu et~al.(2017)Yu, Kulkarni, Lee, and Kim}]{yu2017syllable}
Seunghak Yu, Nilesh Kulkarni, Haejun Lee, and Jihie Kim. 2017.
\newblock Syllable-level neural language model for agglutinative language.
\newblock In {\em Proc. of SCLeM@EMNLP\/}.

\bibitem[{Zaremba et~al.(2014)Zaremba, Sutskever, and
  Vinyals}]{zaremba2014recurrent}
Wojciech Zaremba, Ilya Sutskever, and Oriol Vinyals. 2014.
\newblock Recurrent neural network regularization.
\newblock {\em arXiv preprint arXiv:1409.2329\/} .

\end{thebibliography}
\bibliographystyle{acl_natbib}

\appendix
\section{Optimization}\label{opt}
Training the models involves minimizing the negative log-likelihood over the corpus $w_{1:K}$:
$$\textstyle
-\sum_{k=1}^T\log\Pr(w_k|w_{1:k-1})\longrightarrow\min,\label{nll}
$$
by truncated BPTT \cite{werbos1990backpropagation}. We backpropagate for 35 time steps  using stochastic gradient descent where the learning rate is initially set to 
\begin{itemize}
\item 1.0 in small word-level models,
\item 0.5 in small and medium {\it CharCNN}, medium {\it SylConcat} (SS, SS+RW) models,
\item 0.7 in all other models,
\end{itemize}
and start decaying it with a constant rate after a certain epoch. This is 5 and 10 for the small word-level and all other networks respectively except {\it CharCNN}, for which it is 12. The decay rate is 0.9. The initial values for learning rates were tuned as follows: for each model we start with 1.0 and decrease it by 0.1 until there is convergence at the very first epoch. We use a batch size of 20. We train for 70 epochs. Parameters of the models are randomly initialized uniformly in $[-0.1, 0.1]$ and in $[-0.05, 0.05]$ for the small and medium networks, except the forget bias of the word-level LSTM, which is initialized to $1$, and the transform bias of the highway layer, which is initialized to values around $-2$. For regularization we use a variant of variational dropout \cite{gal2016theoretically} proposed by \citeauthor{inan2016tying} \shortcite{inan2016tying}. For PTB, the dropout rates are 0.3 and 0.5 for the small and medium models. For Wikitext-2, the dropout rates are 0.2 and 0.4 for the small and medium models. We clip the norm of the gradients (normalized by minibatch size) at 5.

For non-English small-sized data sets (Data-S) we use the same hyperparameters as for PTB. To speed up training on non-English medium-sized data (Data-M) we use a batch size of 100 and sampled softmax \cite{DBLP:journals/corr/JeanCMB14} with the number of samples equal to 20\% of the vocabulary size \cite{DBLP:conf/acl/ChenGA16}. 

\section{Non-English corpora statistics and model sizes}\label{sizes}
The non-English small and medium data comes from the 2013 ACL Workshop on Machine Translation\footnote{\url{http://www.statmt.org/wmt13/translation-task.html}} with pre-processing per  \citet{botha2014compositional}. Corpora statistics is provided in Table~\ref{nonenglish_stats}.
\begin{table}
\begin{center}
\begin{tabular}{l l r r r r}
\hline
 & Data set & \multicolumn{1}{c}{$T$} & \multicolumn{1}{c}{$|\mathcal{W}|$} & \multicolumn{1}{c}{$|\mathcal{M}|$}\\
 \hline
\parbox[t]{1mm}{\multirow{5}{*}{\rotatebox[origin=c]{90}{Small}}}
 & French (FR)  & 1M & 25K & 6K \\
 & Spanish (ES) & 1M & 27K & 7K\\
 & German (DE)  & 1M & 37K & 8K\\
 & Czech (CS)   & 1M & 46K & 10K\\
 & Russian (RU) & 1M & 62K & 12K\\
\hline
\parbox[t]{1mm}{\multirow{5}{*}{\rotatebox[origin=c]{90}{Medium}}}
 & French (FR)  & 57M & 137K & 26K\\
 & Spanish (ES) & 56M & 152K & 26K\\
 & German (DE)  & 51M & 339K & 39K\\
 & Czech (CS)   & 17M & 206K & 34K\\
 & Russian (RU) & 25M & 497K & 56K\\
\hline
\end{tabular}
\end{center}
\vspace{-10pt}\caption{Non-English corpora statistics. $T=$ number of tokens in training set; $|\mathcal{W}|=$ word vocabulary size; $|\mathcal{M}|=$ morph vocabulary size.}
\label{nonenglish_stats}
\end{table}

Model sizes for \textit{Word+RE} and \textit{MorphSum+RE+RW}, which were evaluated on non-English data sets, are given in Table~\ref{model_sizes}. \textit{MorphSum+RE+RW} requires 45\%--87\% less parameters than \textit{Word+RE}.
\begin{table}
\setlength{\tabcolsep}{5pt}
\begin{center}
\begin{small}
\begin{tabular}{l l r r r r r r}
\hline
& {Model} & FR & ES & DE & CS & RU\\
\hline
\parbox[t]{0.5mm}{\multirow{3}{*}{\rotatebox[origin=c]{90}{Small}}} & Word + RE & 5.6 & 6.1 & 8.0 & 10.0 & 13.4 & \parbox[t]{0.5mm}{\multirow{6}{*}{\rotatebox[origin=c]{90}{Data-S}}} \\
& MorphSum & \multirow{2}{*}{2.1} & \multirow{2}{*}{2.3} & \multirow{2}{*}{2.5} & \multirow{2}{*}{2.8} & \multirow{2}{*}{3.2} & \\
&  + RE + RW\\
\cline{1-7}
\parbox[t]{1mm}{\multirow{3}{*}{\rotatebox[origin=c]{90}{Medium}}} & Word + RE  & 23.0 & 24.3 & 30.6 & 36.9 & 48.0 &  \\
& MorphSum & \multirow{2}{*}{12.6} & \multirow{2}{*}{13.2} & \multirow{2}{*}{13.8} & \multirow{2}{*}{15.0} & \multirow{2}{*}{16.1} & \\
& + RE + RW \\
\hline
\parbox[t]{1mm}{\multirow{3}{*}{\rotatebox[origin=c]{90}{Small}}} & Word + RE & 28.2 & 31.2 & 68.8 & 42.0 & 100.6 & \parbox[t]{0.5mm}{\multirow{3}{*}{\rotatebox[origin=c]{90}{Data-M}}} \\
& MorphSum & \multirow{2}{*}{6.2} & \multirow{2}{*}{6.1} & \multirow{2}{*}{8.9} & \multirow{2}{*}{7.7} & \multirow{2}{*}{12.6} & \\
& + RE + RW \\
\hline
\end{tabular}
\end{small}
\end{center}
\vspace{-10pt}\caption{Model sizes in millions of trainable parameters.}
\label{model_sizes}
\end{table}

\end{document}